\documentclass{article}

     \PassOptionsToPackage{numbers,compress}{natbib}

     \usepackage[preprint]{neurips_2020}

\usepackage[utf8]{inputenc} 
\usepackage[T1]{fontenc}    
\PassOptionsToPackage{hyphens}{url}\usepackage{hyperref}       
\usepackage{booktabs}       
\usepackage{amsfonts}       
\usepackage{nicefrac}       
\usepackage{microtype}      
\usepackage{xspace}
\usepackage{float}
\usepackage{multirow}
\usepackage{amsmath}
\usepackage{graphicx,caption}
\newcommand*{\eg}{e.g.\@\xspace}
\newcommand*{\ie}{i.e.\@\xspace}
\newcommand*{\wrt}{w.r.t.\@\xspace}
\usepackage[inline]{enumitem}
\usepackage{subcaption}
\usepackage{wrapfig}
\usepackage{listings}
\usepackage{tikz}
\usetikzlibrary{bayesnet}
\usepackage{color, colortbl}
\definecolor{Gray}{gray}{0.9}
\newcolumntype{g}{>{\columncolor{Gray}}r}
\newcommand{\midsepremove}{\aboverulesep = 0mm}
\newcommand{\midsepdefault}{\aboverulesep = 0.605mm \belowrulesep = 0.984mm}

\title{Learning Invariances for Interpretability using Supervised VAE}

\author{%
  An-phi Nguyen\\
  IBM Research Europe, ETH Z\"urich\\
  \texttt{uye@zurich.ibm.com} \\
  \And
  Mar\'ia Rodr\'iguez Mart\'inez\\
  IBM Research Europe\\
  \texttt{mrm@zurich.ibm.com} \\
}

\begin{document}

\maketitle

\begin{abstract}
  We propose to learn model invariances as a means of interpreting a model. This is motivated by a reverse engineering principle. If we understand a problem, we may introduce inductive biases in our model in the form of invariances. Conversely, when interpreting a complex supervised model, we can study its invariances to understand how that model solves a problem. To this end we propose a supervised form of variational auto-encoders (VAEs). Crucially, only a subset of the dimensions in the latent space contributes to the supervised task, allowing the remaining dimensions to act as nuisance parameters. By sampling solely the nuisance dimensions, we are able to generate samples that have undergone transformations that leave the classification unchanged, revealing the invariances of the model.
  Our experimental results show the capability of our proposed model both in terms of classification, and generation of invariantly transformed samples. Finally we show how combining our model with feature attribution methods it is possible to reach a more fine-grained understanding about the decision process of the model.
\end{abstract}

\section{Introduction and Motivation}

The field of interpretable machine learning aims at augmenting the predictions of data-driven models with a human-comprehensible explanation of the reasons behind such predictions. Methods of interpretability are classically divided in four main categories~\cite{Molnar2019InterpretableLearning}, not necessarily mutually exclusive: 
\begin{itemize}
    \item intrinsically interpretable methods, such as decision trees~\cite{Breiman1996BaggingPredictors} or self-explaining neural networks (SENN)~\cite{AlvarezMelis2018TowardsNetworks};
    \item feature attribution methods~\cite{Ancona2018TowardsNetworks,Lundberg2017APredictions}, which assign an importance value to each feature according to their contribution to the prediction;
    \item example-based methods, which implicitly describe the model behavior using examples~\cite{Kim2016ExamplesInterpretability}, possibly adversarial or counterfactual~\cite{Dhurandhar2018ExplanationsNegatives}; or
    \item model internals investigation, which are strategies aiming to understand how the information flows through a model, typically a deep learning model~\cite{Kim2018InterpretabilityTCAV}.
\end{itemize}

While different in many aspects, virtually all the previous methods are concerned with identifying the \emph{important factors} for a prediction. 

In this paper we propose to learn the invariant transformations of an inference model as a novel way to explain it. This is motivated by a reverse engineering principle. Whenever we understand a problem, we design our model so that it \emph{ignores factors that are not important} for the prediction. Some classic examples of this principle are translation-invariance of convolutional neural networks (CNNs)~\cite{LeCun1998Gradient-basedRecognition}, and their recent extensions to more general invariances~\cite{Cohen2016GroupNetworks}. 

For interpretability, this means that a viable way to interpret a model, and understand its underlying prediction rules, is to unveil its invariances: if we manage to find which factors the model are not using, we can better identify the important ones.

To this end, in this paper we propose to use a supervised formulation of variational auto-encoders (VAEs)~\cite{Kingma2014Auto-encodingBayes}. Such a formulation will enable us to simultaneously learn a classifier and a generative model for invariances.

\section{Problem Formulation}\label{sec:problem}

In this paper we aim at simultaneously learning a classifier and its invariant transformations. More specifically, we are provided with i.i.d. labeled samples $(\mathbf{x}_i, y_i) \in \mathcal{X} \times \mathcal{Y}$, and the goal is to learn a classification function $f: \mathcal{X} \rightarrow \mathcal{Y}$ and a set of transformation functions, possibly sample-specific, $\mathcal{T}(\mathbf{x})$ such that $f(\mathbf{x}) = f(\tau(\mathbf{x}))$ for all $\tau \in T(\mathbf{x})$ and for all $\mathbf{x} \in \mathcal{X}$.

There are two possible approaches to learn the transformations $\tau \in T(\mathbf{x})$. 
\begin{itemize}
    \item \emph{Explicit parameterization}. \emph{Each} transformation $\tau_\theta$ is parameterized by a vector-valued parameter $\theta$. The class of transformations could be readily interpretable, \eg rotations. Otherwise, the transformations may be a \emph{rich} (in terms of representational power) class of parametrized functions, such as neural networks. In both cases, the task is to then learn a sample-dependent distribution $p(\theta|\mathbf{x})$ over \emph{all the possible values} of the transformation parameter $\theta$.
    \item \emph{Implicit parametrization}. Instead of learning transformations explicitly, a possibility is to learn a model $p_\theta(\mathbf{\tilde{x}}|\mathbf{x})$ to \emph{directly generate} samples such that $f(\mathbf{x}) = f(\mathbf{\tilde{x}})$. In this case we would only have to learn a single vector-valued parameter $\theta$.
\end{itemize}

From an interpretability perspective, in this paper we adopt a problem-agnostic approach. This means that we don't make any assumption about the possible invariances that the model could learn. This excludes the interpretable parametrized transformations as a solution to our problem. Between the \emph{rich} parametrized class of transformations and the implicit approach, we opt for an \emph{implicit approach}. 
This is motivated by scalability. In the former approach the class of transformations may need a \emph{high} number of parameters to actually be representative: learning the  high-dimensional $p(\theta|\mathbf{x})$ may be difficult because of the \emph{curse of dimensionality}~\cite{Scott2015MultivariateVisualization}. In the latter, instead, we only need to learn a single $\theta$ which parametrizes the generative model $p_\theta(\mathbf{\tilde{x}}|\mathbf{x})$.

\subsection{Supervised VAE}

\begin{figure}
\centering
\begin{tikzpicture}

  \node[latent] (z) {$\mathbf{z}$};
  \node[obs, below=of z] (x) {$\mathbf{x}$};

  \edge {z} {x} ; %
  \edge [dashed, bend left] {x} {z} ; %

  \plate {zx} {(x)(z)} {$N$} ;

\end{tikzpicture}
\hspace{0.2\textwidth}
\begin{tikzpicture}

  \node[latent] (z) {$\mathbf{z}$};
  \node[obs, below=of z, xshift=-1.0cm] (y) {$y$};
  \node[obs, below=of z, xshift=1.0cm] (x) {$\mathbf{x}$};

  \edge {z} {y} ; %
  \edge {z} {x} ; %
  \edge [dashed, bend left] {x} {z} ; %
  \edge [dashed, bend right] {y} {z} ; %
  \edge [dashed, bend right] {x} {z} ; %

  \plate {zyx} {(x)(y)(z)} {$N$} ;

\end{tikzpicture}
\hspace{0.1\textwidth}
\begin{tikzpicture}

  \node[latent] (z) {$\mathbf{z}$};
  \node[obs, below=of z, xshift=1.0cm] (x) {$\mathbf{x}$};
  \node[obs, above=of x, xshift=1.0cm] (y) {$y$};

  \edge {z} {x} ; %
  \edge {y} {x} ; %
  \edge [dashed, bend left] {x} {z} ; %
  \edge [dashed, bend right] {x} {y} ; %
  \edge [dashed] {y} {z} ; %

  \plate {zyx} {(x)(y)(z)} {$N$} ;

\end{tikzpicture}
\caption{Comparison between a traditional unsupervised VAE (left), our proposed Supervised VAE (center), and the Semi-Supervised VAE by \citet{Kingma2014Semi-SupervisedModels} (right). Solid lines denote the generative process, while dashed lines the variational approximation. Observed variables are denoted in gray. Note the different dependency structure between our assumption and the assumption by~\citet{Kingma2014Semi-SupervisedModels}.}\label{fig:graph_model_vae}
\end{figure}
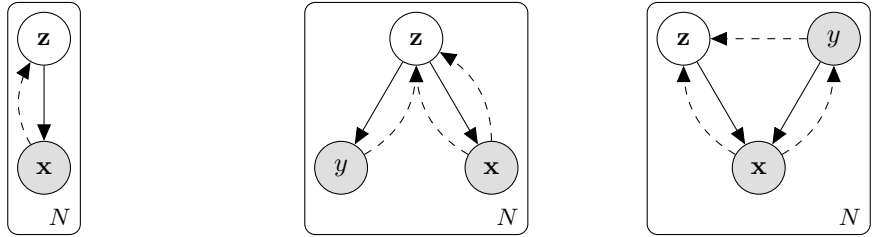

In this paper we propose to use a \emph{supervised generative model} to simultaneously learn a classifier and (implicitly) its invariances. In particular, we propose a modification of \emph{variational auto-encoder} (VAE)~\cite{Kingma2014Auto-encodingBayes}. VAEs are probabilistic models traditionally trained for unsupervised learning by optimization of the \emph{evidence lower bound} (ELBO) $\mathcal{L}(\theta, \phi, \mathbf{x}_i)$:

\begin{equation}\label{eq:vae_elbo}
    \log p (\mathbf{x}_i) \geq \mathbb{E}_{q_\phi} [\log p_\theta (\mathbf{x}_i| \mathbf{z})] - D_{KL}[q_\phi(\mathbf{z}|\mathbf{x}_i)||p(\mathbf{z})] = \mathcal{L}(\theta, \phi, \mathbf{x}_i),
\end{equation}

where the expectation $\mathbb{E}_{q_\phi}$ is taken \wrt the conditional distribution $q_\phi(\mathbf{z}|\mathbf{x})$ of the latent variable $\mathbf{z} \in \mathcal{Z}$ given the observation $\mathbf{x}$, and $D_{KL}[q_\phi(\mathbf{z}|\mathbf{x}_i)||p(\mathbf{z})]$ is the KL divergence between $q_\phi(\mathbf{z}|\mathbf{x})$ and a prior $p(\mathbf{z})$ over the latent space. $p_\theta (\mathbf{x}_i| \mathbf{z})$ and $q_\phi(\mathbf{z}|\mathbf{x})$ are usually referred to as, respectively, probabilistic \emph{decoder} and \emph{encoder}. 

In a supervised setting, among the observed variables we have also the label $y_i$. Figure~\ref{fig:graph_model_vae} compares the probabilistic models implemented by the unsupervised VAE (left) and the supervised VAE for our setting (center).
Eq.~\eqref{eq:vae_elbo} can be easily extended to the supervised case in two ways:

\begin{equation}\label{eq:sup_vae_elbo_discriminative}
    \log p (y_i | \mathbf{x}_i) \geq \mathbb{E}_{q_\phi} [\log p_\theta (y_i| \mathbf{z})] - D_{KL}[q_\phi(\mathbf{z}|\mathbf{x}_i, y_i)||q_\phi(\mathbf{z}|\mathbf{x}_i)],
\end{equation}

\begin{equation}\label{eq:sup_vae_elbo_generative}
    \log p (\mathbf{x}_i, y_i) \geq \mathbb{E}_{q_\phi} [\log p_\theta (\mathbf{x}_i| \mathbf{z})] + \mathbb{E}_{q_\phi} [\log p_\theta (y_i| \mathbf{z})] - D_{KL}[q_\phi(\mathbf{z}|\mathbf{x}_i, y_i)||p(\mathbf{z})],
\end{equation}

While Eq.~\eqref{eq:sup_vae_elbo_generative} is more focused on the generative process, Eq.~\eqref{eq:sup_vae_elbo_discriminative} is more apt to a classification task. This can be seen by analyzing the terms in the different lower bound formulations, as we discuss in the following.

At inference time, for an unseen sample $\mathbf{\hat{x}}$, we would like to classify it by maximization of the posterior
\begin{equation}\label{eq:classifier}
    f(\mathbf{\hat{x}}) = p_\theta(y|\mathbf{\hat{x}}) = \int_\mathcal{Z} p_\theta(y|\mathbf{z})q_\phi(\mathbf{z}|\mathbf{\hat{x}})\mathrm{d}\mathbf{z}
\end{equation}
Only Eq.~\eqref{eq:sup_vae_elbo_discriminative} allows us to learn the distribution $q_\phi(\mathbf{z}|\mathbf{\hat{x}})$. From the decoder in Eq.~\eqref{eq:sup_vae_elbo_generative} it would be technically possible to approximate $q_\phi(\mathbf{z}|\mathbf{\hat{x}}) = \frac{1}{N_i}\sum_i^{N_i} q_\phi(\mathbf{z}|\mathbf{x}_i, y_i)$ \emph{if we had samples} $(\mathbf{x}_i, y_i)$ with $\mathbf{x}_i = \mathbf{\hat{x}}$. However, for an unseen sample, this is not generally true, especially if $\mathcal{X}$ is continuous.

On the other hand, for interpretability purposes, we would like to generate invariantly transformed versions $\mathbf{\tilde{x}}$  of $\mathbf{\hat{x}}$ through the stochastic process
\begin{equation}
    \mathbf{\tilde{x}} \sim p_\theta(\mathbf{\tilde{x}}|\mathbf{z}) \qquad \mathbf{z} \sim q_\phi(\mathbf{z}|\mathbf{\hat{x}})
\end{equation}
Eq.~\eqref{eq:sup_vae_elbo_discriminative}, however, does not provide a way to learn the decoder $p_\theta(\mathbf{\tilde{x}}|\mathbf{z})$.

To tackle both problems, we propose a convex combination of the ELBOs in Eq.~\eqref{eq:sup_vae_elbo_discriminative} and \eqref{eq:sup_vae_elbo_generative}:

\begin{subequations}\label{eq:beta_elbo}
\begin{align}
    \beta \log p (\mathbf{x}_i, y_i) + (1-\beta) \log p (y_i | &\mathbf{x}_i) \geq \mathcal{L}_\beta(\theta, \phi, \mathbf{x}_i, y_i) \\
    \; \mathrm{ with } \quad \mathcal{L}_\beta(\theta, \phi, \mathbf{x}_i, y_i) = \;\; &\beta \mathbb{E}_{q_\phi} [\log p_\theta (\mathbf{x}_i| \mathbf{z})] \label{eq:rec_elbo}\\
    &- \beta D_{KL}[q_\phi(\mathbf{z}|\mathbf{x}_i, y_i)||p(\mathbf{z})])  \label{eq:reg_elbo}\\
     &- (1-\beta)D_{KL}[q_\phi(\mathbf{z}|\mathbf{x}_i, y_i)||q_\phi(\mathbf{z}|\mathbf{x}_i)] \label{eq:suff_elbo}\\ &+\mathbb{E}_{q_\phi} [\log p_\theta (y_i| \mathbf{z})]\label{eq:class_elbo}
\end{align}
\end{subequations}

Similarly to the original VAE, each term in Eq.~\eqref{eq:beta_elbo} plays a distinct role during optimization. Eq.~\eqref{eq:rec_elbo} and \eqref{eq:reg_elbo} are the usual \emph{reconstruction} term and \emph{regularization} term, respectively. Eq.~\eqref{eq:class_elbo} is a (latent-based) \emph{classifier}. We refer to Eq.~\eqref{eq:suff_elbo} as the \emph{sufficiency} term. The reason is that Eq.~\eqref{eq:suff_elbo} ensures that $q_\phi(\mathbf{z}|\mathbf{x}_i)$ is a \emph{sufficient statistic}~\cite{CasellaStatisticalInference} for $q_\phi(\mathbf{z}|\mathbf{x}_i, y_i)$, \ie it ensures that the latent variable $\mathbf{z}$ generated from $\mathbf{x}_i$ contains the same information as if it was jointly generated from $(\mathbf{x}_i, y_i)$.

We note that Eq.~\ref{eq:beta_elbo} is similar in spirit to the $\beta-$VAE~\cite{Higgins2017Beta-VAE:VariationalFramework}. In the traditional formulation of the $\beta-$VAE, the parameter $\beta$ is used to trade-off between the \emph{disentanglement} of the latent dimensions and the reconstruction. In our case, the trade-off is between the generative process and the classification performance, in particular the sufficiency term.

As a final remark, we point out that, while the probabilistic formulation implicitly balances the terms of the ELBO, its optimization may be difficult in practice, depending on the dataset. Following the same strategy as \citet{Kingma2014Semi-SupervisedModels}, we introduce a weight $\alpha > 0$ for the latent classifier in Eq.~\eqref{eq:class_elbo}:  $\mathbb{E}_{q_\phi} [\log p_\theta (y_i| \mathbf{z})] \rightarrow \alpha\mathbb{E}_{q_\phi} [\log p_\theta (y_i| \mathbf{z})]$.

\subsection{Interpretable disentanglement of classifier and invariances}\label{sec:interpretable}

For interpretability purposes, we want to generate invariantly transformed samples by navigating the latent space $\mathcal{Z}$. To do this we need two disentangled sets of dimensions: one that can be used for classification, and another for generation of samples that are invariant to the classification.

We have previously seen that the $\beta$ factor controls the trade-off between the generative process and the classification. In particular, one may think that a higher value of $\beta$ may help promoting disentanglement of the latent dimensions. Unfortunately, \citet{Locatello2019ChallengingRepresentations} have shown that an unsupervised setting cannot in general guarantee that disentangled factors can be identified, and some form of inductive bias or supervision is needed.

Despite having a supervised setting, a naive implementation of Eq.~\eqref{eq:beta_elbo} does not help promoting \emph{interpretable} disentanglement since there is nothing in the formulation indicating which dimensions will be used for classification. This can be easily solved by \emph{a-priori} separating the latent dimensions in two sets $\mathbf{z} = (\mathbf{z}_1, \mathbf{z}_2) \in \mathcal{Z}_1 \times \mathcal{Z}_2$. We can now constrain the classification to use only one set. Eq.~\eqref{eq:classifier} and Eq.~\eqref{eq:class_elbo} can then be respectively rewritten as
\begin{equation}\label{eq:subset_classifier}
    f(\mathbf{\hat{x}}) = \int_{\mathcal{Z}_1} p_\theta(y|\mathbf{z}_1)q_\phi(\mathbf{z}_1|\mathbf{\hat{x}})\mathrm{d}\mathbf{z}_1 \quad \mathrm{and} \quad
    \mathbb{E}_{q_\phi} [\log p_\theta (y_i| \mathbf{z}_1)]
\end{equation}

The remaining dimensions $\mathbf{z}_2$ are now effectively nuisance parameters \wrt the classification and are free to learn how to generate invariant samples. The reconstruction $p_\theta (\mathbf{x}| \mathbf{z})$ is still conditioned on the \emph{entire} latent space.



\section{Experimental Results}

Before demonstrating how the Supervised VAE (SVAE) can be used for interpretability, we quantitatively compare our model to the Semi-Supervised VAE (SemiVAE) proposed by ~\citet{Kingma2014Semi-SupervisedModels} in terms of classification accuracy and ability to generate invariances. Note that the semi-supervised setting chosen by the authors lead them to a different modeling choice, which is shown in Figure~\ref{fig:graph_model_vae} (right). Nonetheless, their model can be used directly in a supervised setting. We focus our analysis on what the authors refer to as the M2 model. The ELBO for this model is:
\begin{equation}\label{eq:semivae_elbo}
\begin{split}
    &\mathcal{L}^{\mathrm{SemiVAE}}(\mathbf{x}_i, y_i) = \mathcal{U}(\mathbf{x}_i, y_i) + \sum_{c \in \mathcal{Y}}q_{\phi}(c|\mathbf{x}_i)\mathcal{U}(\mathbf{x}_i, c) + H(q_{\phi}(y|\mathbf{x}_i)) + \frac{\alpha}{N} \sum_{k}^{N}\log q_{\phi}(y_k|\mathbf{x}_k)\\
    &\mathrm{with} \qquad
    \mathcal{U}(\mathbf{x}_i, y_i)= \mathbb{E}_{q_\phi(\mathbf{z}|\mathbf{x}_i, y_i)} [\log p_\theta (\mathbf{x}_i| y_i, \mathbf{z})] - D_{KL}[q_\phi(\mathbf{z}|\mathbf{x}_i, y_i)||p(\mathbf{z})]
\end{split}
\end{equation}

where $H(\cdot)$ denotes the entropy of a distribution. In the following experiments we do not consider subsequent work based on \cite{Kingma2014Semi-SupervisedModels}, such as \cite{Siddharth2017LearningModels,Xu2016VariationalClassification}, since they are mostly an extension of the original semi-supervised framework to data with different structural assumptions, which is beyond the scope of this paper.

To keep the comparison fair, we use the same architectures for encoders and decoders, as well as the same latent space distribution. Details about architectures and training, and further results are provided in the supplementary material.

In Table~\ref{tab:beta_vs_elbo} we report the reconstruction ELBO and test accuracy for the two models. Note that the architectures for the encoders and the decoders were not optimized for accuracy since here we are mainly concerned with studying the models as supervised \emph{generative} models. 

The results show that our model performs better than the SemiVAE for classification, especially for harder tasks. This can be explained by the fact that the SemiVAE was specifically designed for partially observed labels $y$. In fact, the last term $\frac{1}{N} \sum_{k}^{N}(-\log q_{\phi}(y_k|\mathbf{x}_k))$ in Eq.~\eqref{eq:semivae_elbo} is an additional term that was added (using different variational assumptions) to partially address the classification issues of the original semi-supervised formulation. We refer the readers to~\cite{Kingma2014Semi-SupervisedModels} for further details.

Our model introduces one further parameter $\beta$ to balance the generative process and the classification. For easy tasks, $\beta$ does not seem to influence the classification accuracy. However, it still plays an important role for controlling the reconstruction performance. While in the SemiVAE $\alpha$ can be used to trade-off reconstruction and classification, in the SVAE, the presence of the additional parameter $\beta$ seem to offer better control on the reconstruction performance with reduced impacts on the classification accuracy. Note, however, that the results on CIFAR10 suggest that the trade-off between reconstruction and classification may become more pronounced with the task difficulty.

\midsepremove
\begin{table}
\centering
\caption{Comparison of our proposed Supervised VAE (SVAE) and the Semi-Supervised VAE (SemiVAE):  reconstruction term (R) and test accuracy (\%). For both the models results are reported for two different values of $\alpha$. For MNIST~\cite{LeCun1998Gradient-basedRecognition}, SVHN~\cite{Netzer2011ReadingLearning}, and CIFAR10~\cite{Krizhevsky2009LearningImages}, the values of $(\alpha_1, \alpha_2)$ are, respectively, $(60,6000)$, $(7000,14000)$, $(6000,12000)$. For SVAE we report also values for $\beta$. Averages over 5 runs.} \label{tab:beta_vs_elbo}

\begin{tabular}{lggggggggg}
\toprule
\rowcolor{white}
&   & \multicolumn{4}{c}{SVAE} & \multicolumn{2}{c}{SemiVAE} \\
\cmidrule(l{2pt}r{2pt}){3-6}\cmidrule(l{2pt}r{2pt}){7-8}
\rowcolor{white}
&   & \multicolumn{2}{c}{$\alpha_1$} & \multicolumn{2}{c}{$\alpha_2$} & \multirow{2}{*}{$\alpha_1$} & \multirow{2}{*}{$\alpha_2$} \\
\rowcolor{white}
\cmidrule(l{2pt}r{2pt}){3-4}\cmidrule(l{2pt}r{2pt}){5-6}
\rowcolor{white}
& $\beta$ &  0.01 &   0.9 &  0.01&   0.9 & &\\
\midrule
\rowcolor{white}
\multirow{2}{*}{MNIST} & R &  -402.06 &   46.33 & -682.51  &   \textbf{60.03} & -461.27 & -428.47\\
& \%  &   97.87 &   96.63 &   98.12  &   \textbf{98.23} & 97.35 &  97.30\\
\midrule
\rowcolor{white}
\multirow{2}{*}{SVHN}  & R &  -1214.95 &   -888.05 & -1098.50  &   \textbf{-761.72} & -905.35 & -820.29\\
& \%  &   \textbf{90.78} &   90.39 &   90.57  &   90.61 & 83.85 & 84.42\\
\midrule
\rowcolor{white}
\multirow{2}{*}{CIFAR10}  & R &  -1858.22 &   \textbf{-1314.71} & -1866.77  &   -1340.66 & -1529.63 & -1407.43\\
& \%  &   \textbf{73.58} &   71.45 &   72.99  &   72.24 & 63.07 & 62.21\\
\bottomrule
\end{tabular}
\end{table}
\midsepdefault

%



\subsubsection{Are invariances really invariances?}\label{sec:invariance_test}

In the previous paragraph, we have shown that the parameter $\beta$ can be used to balance the generative and the discriminative parts of our model. This is important because ultimately, while we want to maintain good accuracy, we also want to be able to generate invariantly transformed samples: even if we can efficiently sample invariances in the latent space $\mathcal{Z}$, if we have poor reconstruction performance, we will not be able to correctly observe the invariances in the original input space.

\begin{figure}
\begin{subfigure}{0.5\textwidth}
    \centering
    \includegraphics[width=\linewidth]{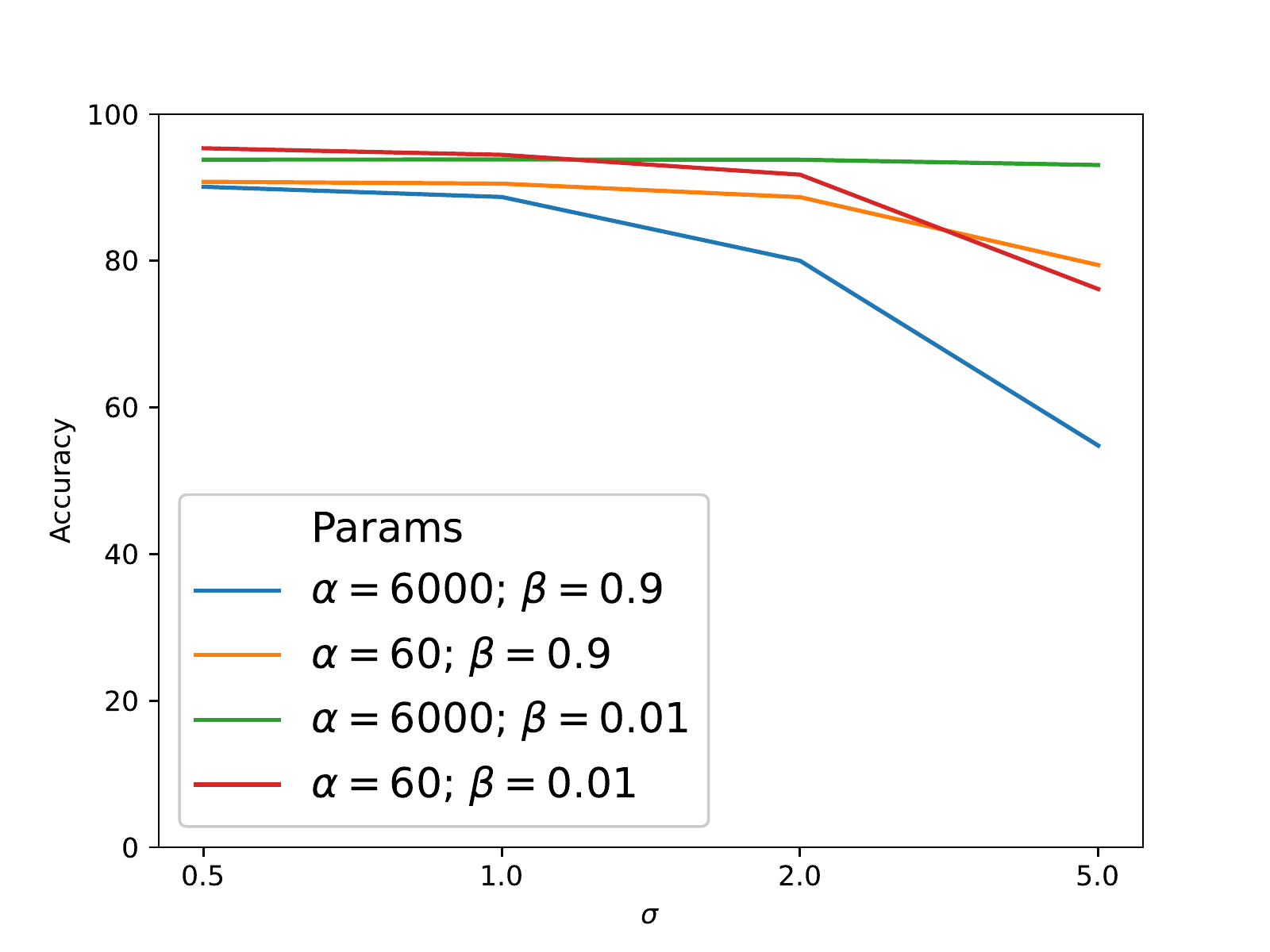}
    \caption{SVAE.}
    \label{fig:svae_invariance_test_our}
\end{subfigure}
\hfill
\begin{subfigure}{0.5\textwidth}
    \centering
    \includegraphics[width=\linewidth]{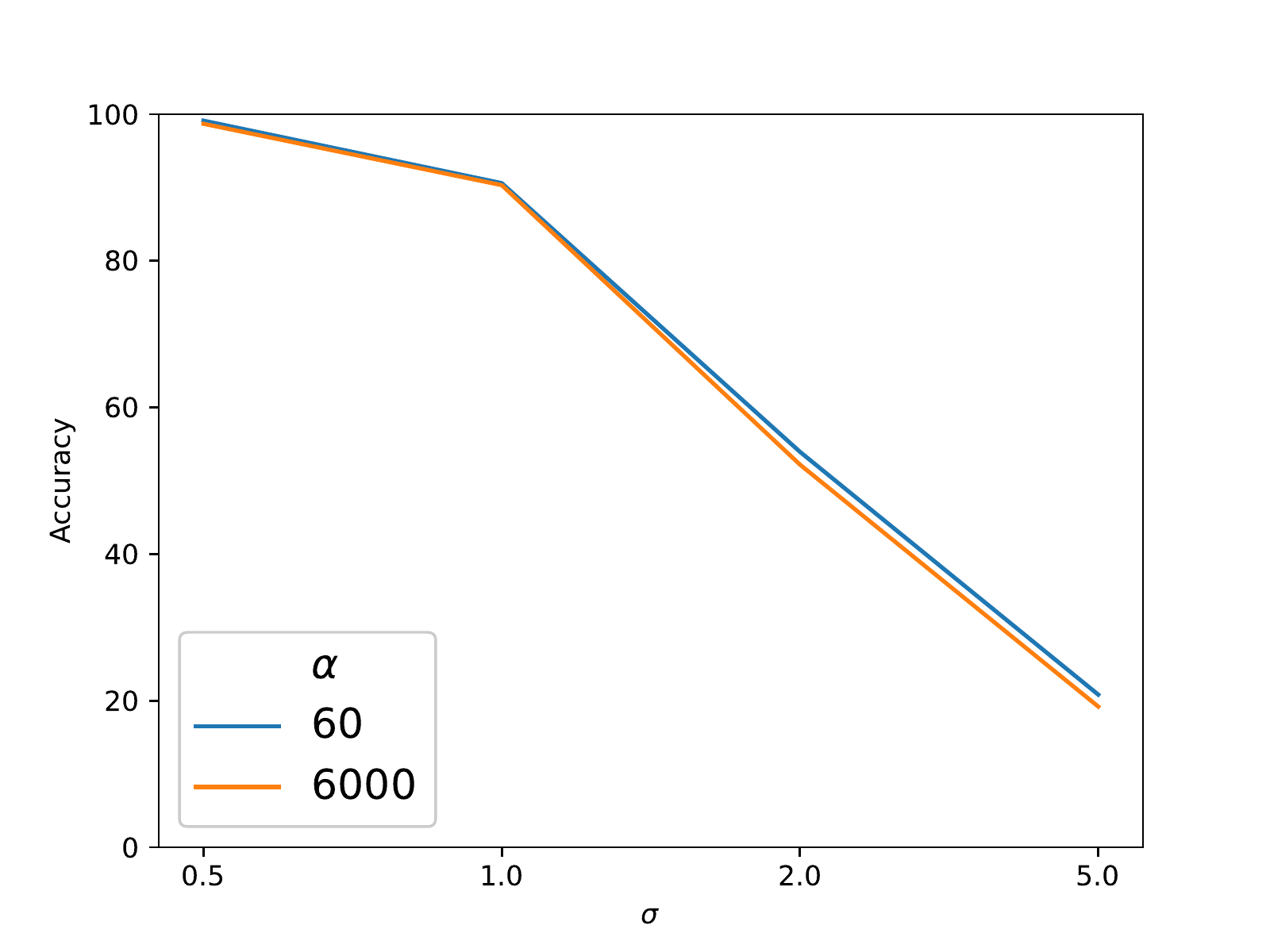}
    \caption{SemiVAE.}
    \label{fig:semivae_invariance_test_kingma}
\end{subfigure}
\caption{Invariance test for SVAE (left) and SemiVAE (right) trained on MNIST. For each parameter set from Table~\ref{tab:beta_vs_elbo}, we plot the percentage of test samples that maintain the same label after being invariantly transformed as a function of the sampling standard deviation $\sigma$.}\label{fig:svae_invariance_test}
\end{figure}

To study how the reconstruction performance relates to the generation of invariantly transformed samples, we perform a simple test. First, we encode test samples to the latent space. Before decoding, we resample the nuisance dimensions according to a zero-centered normal $\mathbf{z}_2 \sim \mathcal{N}(0, \sigma^2)$ for values of $\sigma$ ranging from $0.1$ to $5.0$. We then feed the newly generated samples back into the VAE for classification. Ideally, the predicted label after transformation should not be different from the predicted label for the original sample. Figure~\ref{fig:svae_invariance_test} shows the results for this invariance test for the SVAE and the SemiVAE. 

The results suggest that the SVAE is more robust to changes in the invariant dimensions. The best performing SVAE model ($\alpha=6000$ and $\beta=0.9$) seem to be less invariant for higher values of $\sigma$. This is not too concerning since the enforced prior has unitary standard deviation $\sigma=1.0$. Furthermore, an analysis of the $L_2$ norm of the difference between original and generated images (plot reported in the supplementary) suggests that this model also produces more diverse images increasing the potential of exploring invariances.

\subsection{Interpreting invariances}

In the following, we interpret the results on the MNIST dataset for a SVAE model trained with $\beta=0.9$ and $\alpha=6000$ since, from an interpretability perspective, we are usually more interested in being able to interpret a model with good performance.
Because of the findings from the previous section, in the generation of invariances, we limit ourselves to $\sigma < 2.0$. We further ensure that the generated invariances do maintain the same class of the original sample.

\begin{figure}
\begin{minipage}{0.43\textwidth}
    \centering
    \includegraphics[width=\linewidth]{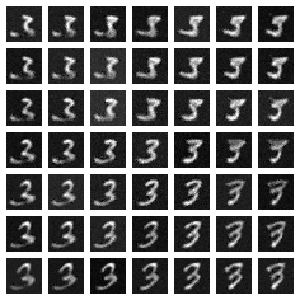}
    \caption{Invariantly transformed samples. The original sample is shown at the center of the grid. The new samples are obtained by crossing two (out of five) randomly picked latent dimensions  with fixed step size .}
    \label{fig:inv_explore}
\end{minipage}
\hfill
\begin{minipage}{0.53\textwidth}
    \centering
    \begin{minipage}{0.66\textwidth}
    \includegraphics[trim=0.4cm 0 6.0cm 0,clip=true,width=\linewidth]{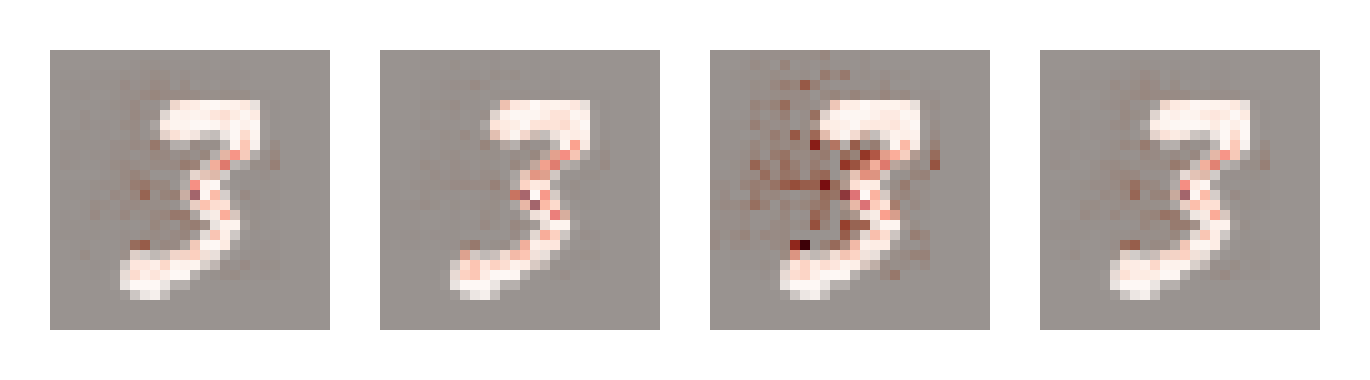}
    \end{minipage}
    
    \begin{minipage}{0.66\textwidth}
    \includegraphics[trim=6.0cm 0 0.4cm 0.5cm,clip=true,width=\linewidth]{figures/vanilla_exp_3_0_1_2_3_4_5_6_7_3_9_3_3_3_3_3.png}
    \end{minipage}
    \caption{Direct application of feature attribution methods to explain the (correctly) predicted class 3: \emph{GradientShap} (top left), \emph{InputXGradient} (top right), \emph{Saliency} (bottom left),and \emph{IntegratedGradients} (bottom right).}
    \label{fig:vanilla_exp}
\end{minipage}
\end{figure}

\begin{figure}
\begin{minipage}[t]{0.48\textwidth}
    \centering
    \includegraphics[trim=5.7cm 0 0 0 ,clip=true,scale=1.25,width=\linewidth]{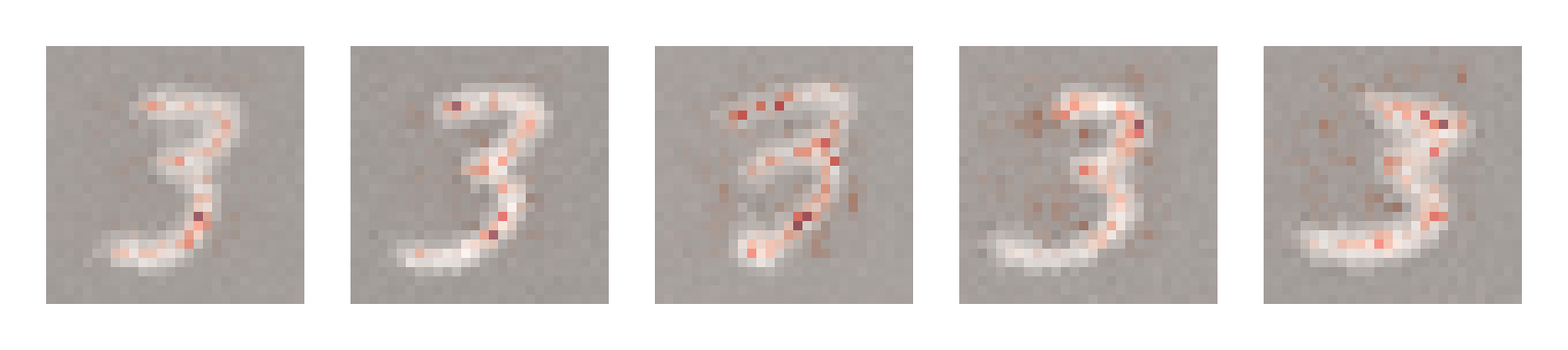}
    \caption{\emph{GradientShap} used to explain Eq.~\eqref{eq:divergence_interpret} with $k=2$ to highlight unimportant factors in 3 generated invariant samples. Upper and lower strokes seem unimportant.}
    \label{fig:unimportant_factors}
\end{minipage}
\hfill
\begin{minipage}[t]{0.48\textwidth}
    \centering
    \includegraphics[trim=5.7cm 0 0 0 ,clip=true,scale=1.25,width=\linewidth]{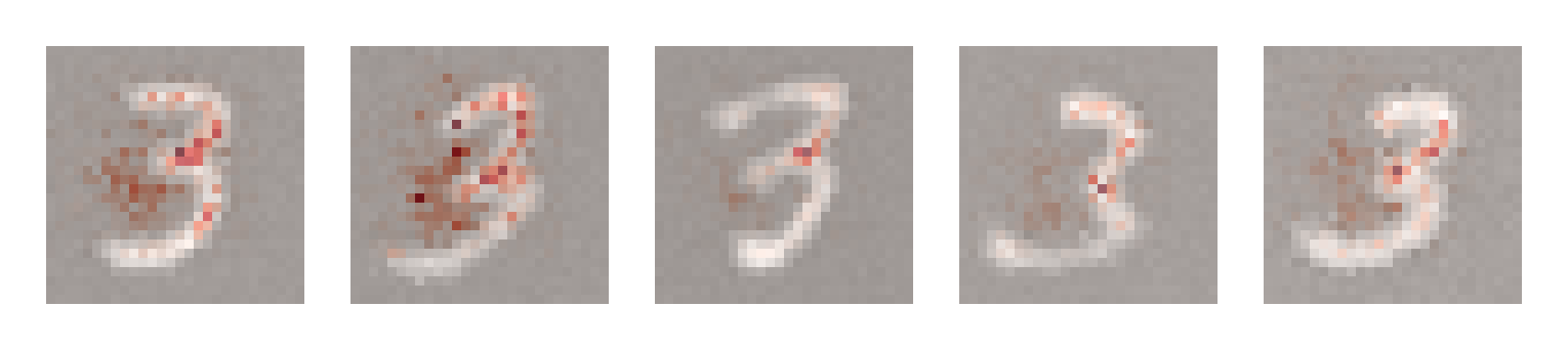}
    \caption{\emph{GradientShap} used to explain Eq.~\eqref{eq:divergence_interpret} with $k=1$ to highlight important factors in 3 generated invariant samples. The crossing point between the upper and lower strokes is important for classifying a digit as 3.}
    \label{fig:direct_comparison_exp}
\end{minipage}
\end{figure}

\begin{figure}
    \centering
    \includegraphics[scale=0.48]{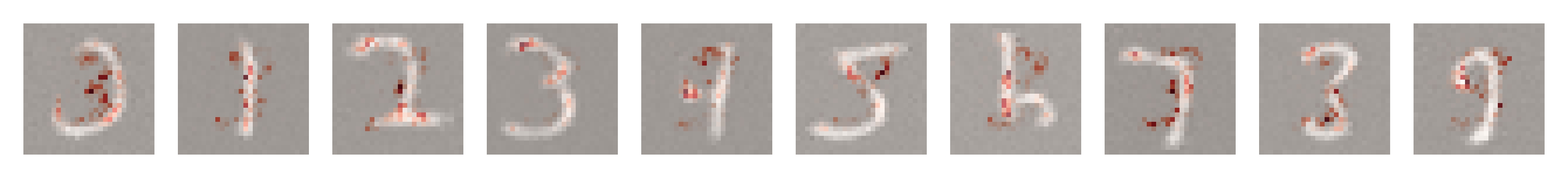}
    \caption{Counterfactual explanations using Eq.~\eqref{eq:divergence_interpret}. The 10 counterfactual examples are generated by individually perturbing the 10 latent dimensions used for classification. The displayed digits go in an increasing fashion from 0 (leftmost) to 9 (rightmost).}
    \label{fig:counter_comparison}
\end{figure}

A first possible way to interpret the (implicitly) learned invariances is by \emph{direct inspection} of the invariantly transformed samples. To this end, it is sufficient to follow the same procedure done for the invariance test (Section~\ref{sec:invariance_test}): first encode the sample we want to interpret, and then generate new samples with modified invariant dimensions. 

Figure~\ref{fig:inv_explore} shows an example of invariantly transformed samples generated by this procedure starting from a sample correctly classified as digit 3. The explored dimensions seem to suggest that, to classify a digit as 3, some style-related features are \emph{not important} (\eg how wide are either the lower or the upper part of the digit).

\subsubsection{Integration with Feature Attribution Methods}\label{sec:feat_attr}

Direct inspection is possible for tasks and data types that are easy to quickly grasp. This is often not true, especially for high-dimensional data with no easily interpretable semantic, \eg genomic sequences. 

In these cases, we propose to increase the interpretability of our model by integration with feature attribution methods~\cite{Ancona2018TowardsNetworks}. This is not achieved by direct application of the feature methods on our classifier, but by \emph{comparison} of these transformed samples with the original one. More specifically, inspired by~\citet{Eberle2020BuildingModels}, instead of interpreting the classification function in Eq.~\eqref{eq:classifier}, we propose to interpret the divergence
\begin{equation}\label{eq:divergence_interpret}
    d(\mathbf{x}, \mathbf{\tilde{x}}) = D_{KL}(q(\mathbf{z}_k|\mathbf{x})||q(\mathbf{z}_k|\mathbf{\tilde{x}})) \qquad \mathrm{for} \; k \in \{1,2\}
\end{equation}
as function of the transformed sample $\mathbf{\tilde{x}}$, while keeping the original sample $\mathbf{x}$ fixed. $k$ denotes the set of latent dimensions we are considering (Section~\ref{sec:interpretable}). With this strategy, we can now highlight both the unimportant factors ($k=2$) and the important ones ($k=1$) that made the model classify the digit as a 3. 

Figure~\ref{fig:vanilla_exp}, Figure~\ref{fig:unimportant_factors} and  Figure~\ref{fig:direct_comparison_exp} show this principle. Figure~\ref{fig:vanilla_exp} shows the direct application of four feature attribution methods~\cite{Lundberg2017APredictions,Simonyan2013DeepMaps,Kindermans2016InvestigatingNetworks,Sundararajan2017AxiomaticNetworks} to explain the classifier $p(y|\mathbf{x})$. They all provide similar results in that they all highlight the general central area of the digit, in agreement with the direct inspection from Figure~\ref{fig:inv_explore}. Stronger evidence is given in Figure~\ref{fig:unimportant_factors}, where the interpretation of the distance in Eq.~\eqref{eq:divergence_interpret} for $k=2$, suggest that the upper and lower strokes of the digit are effectively unimportant.

If we now interpret the divergence with $k=1$  (Figure~\ref{fig:direct_comparison_exp}) we can arguably obtain more \emph{fine-grained} information compared to Figure~\ref{fig:vanilla_exp}: by comparing the original sample with invariantly transformed ones, we are able to conclude that it is actually the exact point where the higher and lower stroke cross that is important for the classification of a digit as a 3. This is particularly evident in the leftmost sample of Figure~\ref{fig:direct_comparison_exp}: the invariant sample has an extended stroke starting from the crossing point which is not highlighted, suggesting that it is not important for the classification.

Note that a similar analysis could be performed if, instead of using generated samples $\mathbf{\tilde{x}}$, we used other samples from the test set correctly classified as 3. However, we argue that our solution has two main advantages.
\begin{itemize}
    \item Using a generative model we can potentially generate \emph{out-of-distribution} samples. This in turn would help us to identify more general, or \emph{global} (in the interpretability sense~\cite{Arya2019OneTechniques}), invariances of the classifier.
    \item Invariantly transformed samples are by definition (Section~\ref{sec:problem}) those samples such that the distance in Eq.~\eqref{eq:divergence_interpret} is zero. In general, a random (but still with the label of interest) sample from the test set may have non-zero distance which we argue may give uninteresting information. Consider two samples (both classified as 3) for which $d(\mathbf{x}, \mathbf{\tilde{x}}) > 0$. This could be given, for example, by the fact that one of the samples has a slightly higher chance of being classified as a 5, rather than an 8 (while still being classified correctly as a 3). This would mean that the question we are actually asking when interpreting this distance is ``Why this 3 looks more like a 5 than an 8 compared to other?''.
\end{itemize}

\subsubsection{Counterfactual explanations}\label{sec:counterfactual}

Conveniently, Eq.~\eqref{eq:divergence_interpret} is not restricted to providing direct evidence. If, instead of generating invariances we generated samples for other classes (by perturbing the dimensions $\mathbf{z}_1$ used by the classifier), we are able to produce counterfactual evidence, \ie evidence that shows why a digit was \emph{not} classified as a 3~\cite{Dhurandhar2018ExplanationsNegatives}. This is shown in Figure~\ref{fig:counter_comparison}. For example, if we observe the generated digit that is classified as 0, we notice that the main missing feature for classifying it as a 3 is the missing crossing point between the upper and lower stroke.

Note that our methodology has an advantage compared to other counterfactual methods  \cite{Dhurandhar2018ExplanationsNegatives,Wachter2017CounterfactualGDPR}. These methods, in order to find a counterfactual example, have to solve an optimization problem \emph{for each sample}. In our case, however, we can readily generate counterfactuals by sampling the latent space, making our strategy more efficient and scalable.

\section{Related Work}

Our work positions itself at the intersection of interpretability and generative modeling. In particular, we propose to learn invariances of a predictive model as a way to explain it.

As a possible approach to implement this idea, we propose to use a supervised formulation of variational auto-encoders~\cite{Kingma2014Auto-encodingBayes} in order to be able to both perform a supervised task and generate invariant samples. Closely related to our formulation is the work from~\citet{Kingma2014Semi-SupervisedModels} which we included in our experiments as a baseline. Subsequent work, such as \cite{Siddharth2017LearningModels,Xu2016VariationalClassification}, present an extension to different structural assumptions for the prior and/or the data. While these prior works were introduced in the context of semi-supervised learning, they can be used in a full supervised setting as well. The main difference between those formulations and ours is the underlying graphical model. Previous work assumes the label $y$ to take part in the generation of the samples $\mathbf{x}$, while we assume both $y$ and $\mathbf{x}$ to be independently generated by the latent code $\mathbf{z}$. The former might be convenient from a disentanglement perspective (since the marginals of $y$ and $\mathbf{z}$ are independent). However, the latent space dimensionality would grow with the number of classes (and tasks for multi-task scenarios~\cite{Caruana1998MultitaskLearning}). 

From an interpretability perspective, we offer a different way of generating explanations. In terms of explanation presentation, however, our current work is similar to example/prototype-based methods~\cite{Kim2016ExamplesInterpretability,Gurumoorthy2017EfficientWeights,Ghorbani2019TowardsExplanations}: to show the invariances, we show examples of invariantly transformed samples. 
Another way to leverage the invariances learned by the model is by integrating feature attribution methods. Our experiments show how we can improve on the explanations provided by feature attribution methods achieving \emph{more specific information} about classification rules. Using the same strategy we are able to also generate counterfactual examples, similarly to the work of \citet{Dhurandhar2018ExplanationsNegatives}. A clear advantage of our method is computational as we discussed in Section~\ref{sec:counterfactual}.

\section{Conclusion and Future Work}

In this paper we present a novel approach towards interpretability, \ie learning invariances of a predictive model: by finding what the predictive model does not care about, we are able to better pinpoint what it does care about.

To this end, we propose a fully-supervised formulation of variational auto-encoders. We empirically compare our model against an established (semi-)supervised generative model, and show how it is more apt to classification tasks, while being able to generate invariances. We then show how these invariances can be used for interpretability, in both a direct way and by leveraging feature attribution methods, ultimately obtaining more detailed explanations. While in this work we showcased our method on benchmark image datasets, we foresee more benefits in applications with high-dimensional data without clear semantic, \eg long genomic sequences. In these cases, similarly to Figure~\ref{fig:vanilla_exp}, classic methods may provide too noisy explanations. 

Other future research directions include the exploration of different ways of learning and understanding invariances, such as the explicit parametrization as discussed in Section~\ref{sec:problem}. One solution to learning the distribution over the parameters $\theta$ is to leverage a bayesian formulation and methods able to update and sample the posterior distribution of the parameters during training, \eg~\cite{Welling2011BayesianDynamics}.

\section*{Broader Impact}

Machine learning models are increasingly deployed in many scenarios. In some scenarios, the model may be required to output a decision that could have potentially deleterious consequences to different members of the society, \eg \cite{AmazonReuters,Ledford2019MillionsAlgorithms}. Interpretable machine learning is a possible way to tackle these issues by augmenting algorithmic decisions with explanations. In this context, our work wishes to provide an \emph{additional} way to generate explanations: the more tools we have to create explanations, the higher are the chances to find problems in our algorithms. As any (generative) machine learning model, our proposed variational auto-encoder could either be manipulated (\eg by adversarial examples~\cite{Xu2020AdversarialReview}) or be used for generation of misleading explanations. We do not however see any potential negative impact specific to our model or our proposed way of generating explanations.

\bibliographystyle{unsrtnat}
\bibliography{references}

\end{document}